\documentclass[conference]{IEEEtran}

\usepackage{cite}  
\usepackage{url}  
\usepackage{graphicx}  
\usepackage{multirow}
\usepackage{algorithmicx}
\usepackage{algpseudocode}
\usepackage{algorithm}
\usepackage{xcolor}
\usepackage{comment}
\usepackage{caption}
\usepackage{subcaption}
\usepackage[export]{adjustbox}

\graphicspath{ {./images/} }

\begin{document}

\title{Exploring Adversarial Examples\\in Malware Detection}

\author{
\IEEEauthorblockN{Octavian Suciu}
\IEEEauthorblockA{\textit{University of Maryland, College Park} \\
osuciu@umiacs.umd.edu}
\and
\IEEEauthorblockN{Scott E. Coull}
\IEEEauthorblockA{\textit{FireEye, Inc.} \\
scott.coull@fireeye.com}
\and
\IEEEauthorblockN{Jeffrey Johns}
\IEEEauthorblockA{\textit{FireEye, Inc.} \\
jeffrey.johns@fireeye.com}
}

\maketitle

\begin{abstract}

The convolutional neural network (CNN) architecture is increasingly being applied to new domains, such as malware detection, where it is able to learn malicious behavior from raw bytes extracted from executables.
These architectures reach impressive performance with no feature engineering effort involved, but their robustness against active attackers is yet to be understood. 
Such malware detectors could face a new attack vector in the form of adversarial interference with the classification model. 
Existing evasion attacks intended to cause misclassification on test-time instances, which have been extensively studied for image classifiers, are not applicable because of the input semantics that prevents arbitrary changes to the binaries.
This paper explores the area of adversarial examples for malware detection.
By training an existing model on a production-scale dataset, we show that some previous attacks are less effective than initially reported, while simultaneously highlighting  
architectural weaknesses that facilitate new attack strategies for malware classification.
Finally, we explore how generalizable different attack strategies are, the trade-offs when aiming to increase their effectiveness, and the transferability of single-step attacks.

\end{abstract}

\section{Introduction}
\label{sec:intro}

The popularity of convolutional neural network (CNN) classifiers has lead to their adoption in fields which have been historically adversarial, such as malware detection~\cite{raff2017malware,krvcal2018deep}.
Recent advances in adversarial machine learning have highlighted weaknesses of classifiers when faced with adversarial samples.
One such class of attacks is evasion~\cite{biggio2013evasion}, which acts on test-time instances.
The instances, also called adversarial examples, are modified by the attacker such that they are misclassified by the victim classifier even though they still resemble their original representation. 
State-of-the-art attacks focus mainly on image classifiers~\cite{szegedy2013intriguing,goodfellow6572explaining,papernot2017practical,carlini2017towards}, where attacks add small perturbations to input pixels that lead to a large shift in the victim classifier feature space, potentially shifting it across the classification decision boundary. 
The perturbations do not change the semantics of the image as a human oracle easily identifies the original label associated with the image.

In the context of malware detection, adversarial examples could represent an additional attack vector for an attacker determined to evade such a system.
However, domain-specific challenges limit the applicability of existing attacks designed against image classifiers on this task.
First, the strict semantics of binary files disallows arbitrary perturbations in the input space. 
This is because there is a structural interdependence between adjacent bytes, and any change to a byte value could potentially break the functionality of the executable.
Second, limited availability of representative datasets or robust public models limits the generality of existing studies.
Existing attacks~\cite{kolosnjaji2018adversarial,kreuk2018deceiving} use victim models trained on very small datasets, and make various assumptions regarding their strategies.  
Therefore, the generalization effectiveness across production-scale models and the trade-offs between various proposed strategies is yet to be evaluated.

This paper sheds light on the generalization property of adversarial examples against CNN-based malware detectors.
By training on a production-scale dataset of 12.5 million binaries, we are able to observe interesting properties of adversarial attacks, showing that their effectiveness could be misestimated when small datasets are used for training, and that single-step attacks are more effective against robust models trained on larger datasets. 

Our contributions are as follows:
\begin{itemize}
  \item We measure the generalization property of adversarial attacks across datasets, highlighting
  common properties and trade-offs between various strategies.
  \item We unearth an architectural weakness of a published CNN architecture that facilitates existing attack strategies~\cite{kolosnjaji2018adversarial,kreuk2018deceiving}.
  \item We investigate the transferability of single-step adversarial examples across models trained on different datasets. 
\end{itemize}

\section{Background}
\label{sec:background}

The CNN architecture has proven to be very successful across popular vision tasks, such as image classification~\cite{he2016deep}.
This lead to an increased adoption in other fields and domains, with one such example being text classification from character-level features~\cite{zhang2015character}, which turns out to be extremely similar to the malware classification problem discussed in this paper. 
In this setting, a natural language document is represented as a sequence of characters, and the CNN is applied on that one-dimensional stream of characters.
The intuition behind this approach is that a CNN is capable of automatically learning complex features, such as words or word sequences, by observing compositions of raw signals extracted from single characters.
This approach also avoids the requirement of defining language semantic rules, and is able to tolerate anomalies in features, such as word misspellings.
The classification pipeline first encodes each character into a fixed-size embedding vector.
The sequence of embeddings acts as input to a set of convolutional layers, intermixed with pooling layers, then followed by fully connected layers. 
The convolutional layers act as receptors, picking particular features from the input instance, while the pooling layers act as filters to down-sample the feature space. 
The fully connected layers act as a non-linear classifier on the internal feature representation of instances.

\subsection{CNNs for Malware Classification.}
Similar to this approach, the security community explored the applicability of CNNs to the task of malware detection.
Raff et al.~\cite{raff2017malware} and Kr{\v c}{\' a}l et al.~\cite{krvcal2018deep} use the CNNs on a raw byte representation, whereas Davis and Wolff~\cite{davis2015deep} use it on disassembled functions.
In this work we focus on the raw byte representation.
In an analogy to the text domain, an executable file could be conceptualized as a sequence of bytes that are arranged into higher-level features, such as instructions or functions.
By allowing the classifier to automatically learn features indicative of maliciousness,
this approach avoids the labor-intensive feature engineering process typical of malware classification tasks.
Manual feature engineering proved to be challenging in the past and led to an arms race between antivirus developers and attackers aiming to evade them~\cite{ugarte2015sok}.
However, the robustness of these automatically learned features in the face of evasion is yet to be understood.

\begin{figure}[t]
\centering
\includegraphics[width=.6\columnwidth]{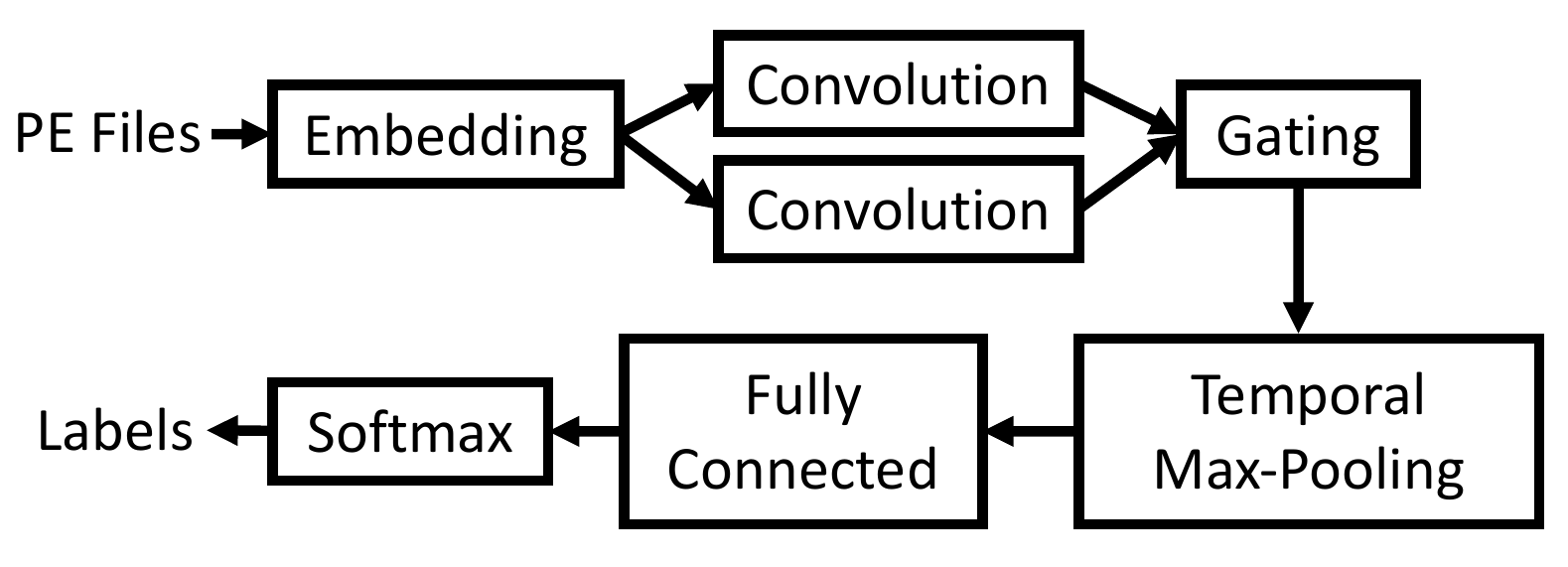}
\caption{Architecture for the MalConv Model.}
\label{fig:malconv}
\end{figure}

In this paper, we explore evasion attacks by focusing on a byte-based convolutional neural network for malware detection, called MalConv~\cite{raff2017malware}, whose architecture is shown in Figure~\ref{fig:malconv}.
MalConv reads up to 2MB of raw byte values from a Portable Executable (PE) file as input, appending a distinguished padding token to files smaller than 2MB and truncating extra bytes from larger files. 
The fixed-length sequences are then transformed into an embedding representation, where each byte is mapped to an 8-dimensional embedding vector.
These embeddings are then passed through a gated convolutional layer, followed by a temporal max-pooling layer, before being classified through a final fully connected layer.
Each convolutional layer uses a kernel size of 500 bytes with a stride of 500 (i.e., non-overlapping windows), and each of the 128 filters is passed through a max-pooling layer.
This results in a unique architectural feature that we will revisit in our results:  each pooled filter is mapped back to a specific 500-byte sequence and there are at most 128 such sequences that 
contribute to the final classification across the entire input.
Their reported results on a testing set of 77,349 samples achieved a Balanced Accuracy of 0.909 and Area Under the Curve (AUC) of 0.982.


\subsection{Adversarial Binaries.}
Unlike evasion attacks on images~\cite{szegedy2013intriguing,goodfellow6572explaining,papernot2017practical,carlini2017towards}, attacks that alter the raw bytes of PE files must maintain the 
syntactic and semantic fidelity of the original file.   
The Portable Executable (PE) standard~\cite{microsoftdocs} defines a fixed structure for these files.
A PE file contains a leading header enclosing file metadata and pointers to the sections of the file, followed by the variable-length sections which contain the actual program code and data.
Changing bytes arbitrarily could break the malicious functionality of the binary or, even worse, prevent it from loading at all.
Therefore, an attacker constrained to static analysis of the binaries has limited leverage on the features they can modify.

Recent work~\cite{kolosnjaji2018adversarial,kreuk2018deceiving} suggests two strategies of addressing these limitations. 
The first one avoids this problem by appending adversarial noise to the end of the binary.
Since the appended adversarial bytes are not within the defined boundaries of the PE file, their existence does not impact the binary's functionality and there are no inherent restrictions on the syntax of 
bytes (i.e., valid instructions and parameters).
The trade-off, however, is that the impact of the appended bytes on the final classification is offset by the features present in the original sample, which remain unchanged.  
As we will see, these attacks take advantage of certain vulnerabilities in position-independent feature detectors present in the MalConv architecture.
The second strategy~\cite{kreuk2018deceiving} seeks to discover regions in the executable that are not mapped to memory and that, upon modification, would not affect the intended behavior. However, the utility of this approach compared to append strategies has not been studied before.
In this paper, we evaluate the comparative effectiveness of the two strategies at scale and highlight their transferability across models, as well as trade-offs that might affect their general applicability.


\subsection{Datasets.}
To evaluate the success of evasion attacks against the MalConv architecture we utilize three datasets.
First, we collected 16.3M PE files from a variety of sources, including VirusTotal, Reversing Labs, and proprietary FireEye data.
The data was used to create a production-quality dataset of 12.5M training samples and 3.8M testing samples, which we refer to as the \textit{Full} dataset.
The corpus contains 2.2M malware samples in the training set, and 1.2M in testing.
The dataset was created from a larger pool of more than 33M samples using a stratified sampling technique based on malware family. Use of stratified sampling ensures uniform coverage over the canonical `types' of binaries present in the dataset, while also limiting bias from certain overrepresented types (e.g.,  popular malware families).  
Second, we utilize the \textit{EMBER} dataset~\cite{2018arXiv180404637A}, which is a publicly available dataset comprised of 1.1M  PE files, out of which 900K are used for training. 
On this dataset, we use the pre-trained MalConv model released with the dataset. 
 In addition, we also created a smaller dataset whose size and distribution is more in line with Kolosnjaji et al.'s evaluation~\cite{kolosnjaji2018adversarial}, which we refer to as the \textit{Mini} dataset.  
The Mini dataset was created by sampling 4,000 goodware and 4,598 malware samples from the Full dataset.  
Note that both datasets follow a strict temporal split where test data was observed strictly later than training data.
We use the Mini dataset in order to explore whether the attack results demonstrated by Kolosnjaji et al. would generalize to a production-quality model, or whether they are artifacts of the dataset properties.

\section{Baseline Performance}
\label{sec:baseline}
To validate our implementation of the MalConv architecture~\cite{raff2017malware}, 
we train the classifier on both the Mini and the Full datasets, leaving out the DeCov regularization addition suggested by the authors.
Our implementation uses a momentum-based optimizer with decay and a batch size of 80 instances.
We train on the Mini dataset for 10 full epochs. 
We also trained the Full dataset for 10 epochs, but stopped the process early due to a small validation loss\footnote{This was also reported in the original MalConv study.}.
To assess and compare the performance of the two models, we test them on the entire Full testing set.
The model trained on the Full dataset achieves an accuracy of 0.89 and an AUC of 0.97, which is similar to the results published in the original MalConv paper.
Unsurprisingly, the Mini model is much less robust, achieving an accuracy of 0.73 and an AUC of 0.82. 
The MalConv model trained on EMBER was reported to achieve 0.99 AUC on the corresponding test set.

\section{Attack Strategies}

We now present the attack strategies used throughout our study and discuss their trade-offs.

\subsection{Append Attacks}
\label{sec:append}

Append-based strategies address the semantic integrity constraints of PE files by appending adversarial noise to the original file.
We start by presenting two attacks first introduced by Kolosnjaji et al.~\cite{kolosnjaji2018adversarial} and evaluated against MalConv, followed by our two strategies intended to evaluate the robustness of the classifier.

\paragraph{Random Append} This attack works by appending byte values sampled from a uniform distribution. 
This baseline attack measures how easily an append attack could offset features derived from the file length, and helps compare the actual adversarial gains from more complex append strategies over random appended noise. 

\paragraph{Gradient Append} The Gradient Append strategy uses the input gradient value to guide the changes in the appended byte values.
The algorithm appends $numBytes$ to the candidate sample and updates their values over $numIter$ iterations or until the victim classifier is evaded.
The gradient of the output with respect to the input layer indicates the direction, in the input space, of the change required to minimize the output, therefore pushing its value towards the benign class.
The representation of all appended bytes is iteratively updated, starting from random values.
However, as the input bytes are mapped to a discrete embedding representation in MalConv, the end-to-end architecture becomes non-differentiable and its input gradient cannot be computed analytically.   
Therefore, this attack uses a heuristic to instead update the embedding vector and discretize it back in the byte space to the closest byte value along the direction of the embedding gradient.  We refer interested readers to the original paper for details of this discretization process~\cite{kolosnjaji2018adversarial}.
The attack requires $numBytes * numIter$ gradient computations and updates to the appended bytes in the worst case, which could be prohibitively expensive for large networks.

\paragraph{Benign Append} 
This strategy allows us to observe the susceptibility of the MalConv architecture, specifically its temporal max-pooling layer, to attacks that reuse benign byte sequences at the end of a file.
The attack takes bytes from the beginning of benign instances and appends them to the end of a malicious instance.
The intuition behind this attack is that leading bytes of a file, and especially the PE headers, are the most influential towards the classification decision \cite{raff2017malware}. 
Therefore, it signals whether the maliciousness of the target could be offset by appending highly influential benign bytes.

\begin{algorithm}
\scriptsize
\caption{The FGM Append attack}
\label{alg:fgmappend}
\begin{algorithmic}[1]
\Function {FGMAppend}{$x_0$, $numBytes$, $\epsilon$}

\State $x_0 \gets \Call{PadRandom}{x_0, numBytes}$
\State $e \gets \Call{GetEmbeddings}{x_0}$ 
\State $e_u \gets \Call{GradientAttack}{e,\epsilon}$
\For{$i$ \textbf{in} $|x_0|...|x_0|+numBytes-1$}
\State $e[i] \gets e_u[i]$ 
\EndFor
\State $x^* \gets \Call{EmbeddingMapping}{e}$
\State \Return $x^*$
\EndFunction

\Function {GradientAttack}{$e$, $\epsilon$}
\State $e_u \gets e -\epsilon*sign(\nabla_l (e))$
\State \Return $e_u$
\EndFunction

\Function {EmbeddingMapping}{$e_x$}
\State $e\gets \Call{Array}{256}$ 
\For{$byte$ \textbf{in} $0...255$}
\State $e[byte] \gets \Call{GetEmbeddings}{byte}$ 
\EndFor
\For{$i$ in $0...|e_x|$}
\State $x^*[i] \gets argmin_{b \in 0...255}(||e_x[i]-e[b]||_2)$ 
\EndFor
\State \Return $x^*$
\EndFunction

\end{algorithmic}
\end{algorithm}

\paragraph{FGM Append} 
Based on the observation that the convergence time of the Gradient Append attack grows linearly with the number of appended bytes, we propose the ``one-shot" FGM Append attack, an adaptation of the Fast Gradient Method (FGM) originally described in~\cite{goodfellow6572explaining}.
The adaptation of the FGM attack to the malware domain was first proposed by Kreuk et al.~\cite{kreuk2018deceiving} in an iterative algorithm intended to generate a small-sized adversarial payload.
In contrast, our attack strategy aims to highlight vulnerabilities of the model as a function of the increasing adversarial leverage.
The pseudocode is described in Algorithm~\ref{alg:fgmappend}.
Our attack starts by appending $numBytes$ random bytes to the original sample $x_0$ and updating them using a policy dictated by FGM.
The attack uses the classification loss $l$ of the output with respect to the target label.
FGM updates each embedding value by a user specified amount $\epsilon$ in a direction that minimizes $l$ on the input, as dictated by the sign of the gradient $\nabla_l$.
While this attack framework is independent of the distance metric used to quantify perturbations, our experiments use $L_\infty$. 
In order to avoid the non-differentiability issue, our attack performs the gradient-based updates of the appended bytes in the embedding space, while mapping the updated value to the closest byte value representation in \textsc{EmbeddingMapping} using the $L_2$ distance metric.
A more sophisticated mapping could be used to ensure that the update remains beneficial towards minimizing the loss. 
However, we empirically observed that the metric choice does not significantly affect the results for our single-step attack.

\subsection{Limitations of Append Strategies}
\label{sec:appendlimitations}

\begin{figure}[t]
\centering
\includegraphics[width=0.7\columnwidth]{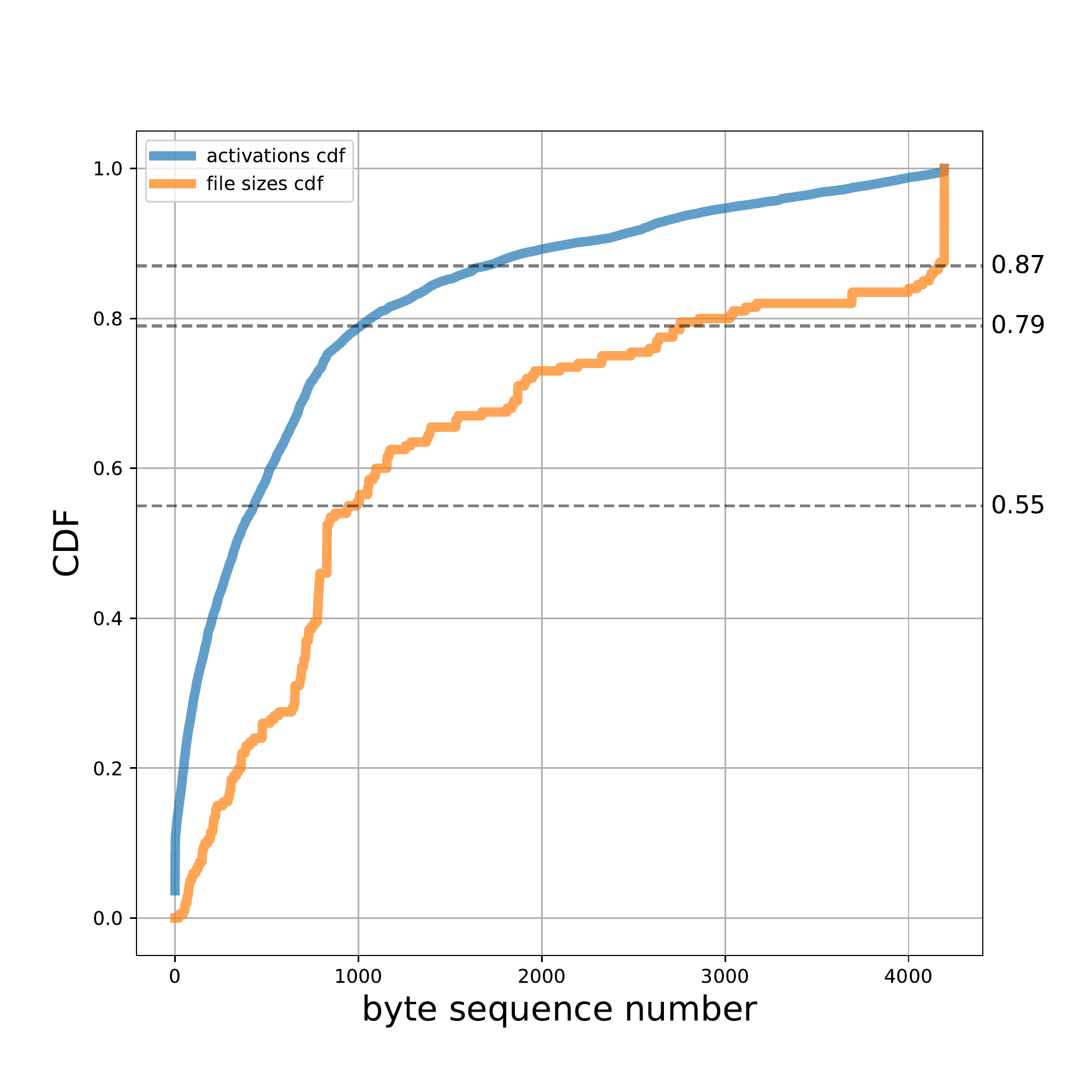}
\caption{CDF of file sizes and activation locations determined by MalConv’s max pooling layer.}
\label{figure:pooling_cdf}
\end{figure}

Besides the inability to append bytes to files that already exceed the model's maximum size (e.g., 2MB for MalConv), append-based attacks can suffer from an additional limitation.
In the MalConv architecture, a PE file is broken into non-overlapping byte sequences of length 500. With a maximum file size of 2MB, that corresponds to at most 4,195 such sequences.
The model uses 128 features, meaning only 128 of the 4,195 sequences can ever be selected.  
In Figure~\ref{figure:pooling_cdf}, we select a random set of 200 candidate malware samples and examine the file size distribution and which of the 4,195 sequences are being selected, on average, by the model.
This shows that, for example, while the first 1,000 sequences (0.5 MB) in binaries correspond to 79\% of the actual features for the classifier, only 55\% of the files are smaller than that.
Additionally, 13\% of the instances cannot be attacked at all because they are larger than the maximum file size for the classifier. 
The result shows not only that appended bytes need to offset a large fraction of the original discriminative features, but also that attacking the byte sequences of these discriminative features directly will likely amplify the attack effectiveness due to their importance.  Driven by this intuition, we proceed to describe an attack strategy that would exploit the existing bytes of binaries \textit{with no side effects on the functionality of the program}.

\subsection{Slack Attacks}
\label{sec:slack}
\paragraph{Slack FGM}  Our strategy defines a set of slack bytes where an attack algorithm is allowed to freely modify bytes in the existing binary without breaking the PE.
Once identified, the slack bytes are then modified using a gradient-based approach.
The \textsc{SlackAttack} function in Algorithm~\ref{alg:slackfgm} highlights the architecture of our attack. The algorithm is independent of the \textsc{SlackIndexes} method employed for extracting slack bytes or the gradient-based method in \textsc{GradientAttack} used to update the bytes. 

\begin{algorithm}
\scriptsize
\caption{The Slack FGM attack}
\label{alg:slackfgm}
\begin{algorithmic}[1]
\Function {SlackAttack}{$x_0$}
\State $m \gets \Call{SlackIndexes}{x_0}$
\State $e \gets \Call{GetEmbeddings}{x_0}$ 
\State $e_u \gets \Call{GradientAttack}{e}$ 
\State $x_u \gets \Call{EmbeddingMapping}{e_u}$
\State $x^* \gets x_0$
\For{$idx$ \textbf{in} $m$}
\State $x^*[idx] \gets x_u[idx]$
\EndFor 
\State \Return $x^*$
\EndFunction

\Function {SlackIndexes}{$x$}
\State $s \gets \Call{GetPESections}{x}$ 
\State $m\gets \Call{Array}{0}$ 
\For{$i$ \textbf{in} $0...|s|$}
\If{$s[i].RawSize > s[i].VirtualSize$}
\State $r_s \gets s[i].RawAddress+s[i].VirtualSize$ 
\State $r_e \gets s[i].RawSize$ 
\For{$idx$ \textbf{in} $r_s...r_e$}
\State $m \gets \Call{Append}{m,idx}$ 
\EndFor
\EndIf
\EndFor
\State \Return $m$
\EndFunction
\end{algorithmic}
\end{algorithm}

In our experiments we use a simple technique that empirically proves to be effective in finding sufficiently large slack regions.
This strategy extracts the gaps between neighboring PE sections of an executable by parsing the executable section header.
The gaps are inserted by the compiler and exist due to misalignments between the virtual addresses and the multipliers over the block sizes on disk.
We compute the size of the gap between consecutive sections in a binary as $RawSize - VirtualSize$, and define its byte start index in the binary by the section's $RawAddress + VirtualSize$.
By combining all the slack regions, \textsc{SlackIndexes} returns a set of indexes over the existing bytes of a file, indicating that they can be modified.
This technique was first mentioned in~\cite{kreuk2018deceiving}.
However, to our knowledge, a systematic evaluation of its effectiveness and the comparison between the slack and append strategies have not been performed before.

Although more complex byte update strategies are possible, potentially accounting for the limited leverage imposed by the slack regions, we use the technique introduced for the FGM Append attack in Algorithm~\ref{alg:fgmappend}, which proved to be effective.
Like in the case of FGM Append, updates are performed on the embeddings of the allowed byte indexes and the updated values are mapped back to the byte values using the $L_2$ distance metric.

\begin{table*}
\scriptsize
\centering
\begin{tabular}{|c||c|c|c||c|c|c||c|c|c|}
\hline
\small \# Append Bytes & \multicolumn{3}{|c||}{\small Random Append} & \multicolumn{3}{|c||}{\small Benign Append} & \multicolumn{3}{|c|}{\small FGM Append} \\ \hline
      & \small Mini & \small EMBER & \small Full & \small Mini & \small EMBER & \small Full & \small Mini & \small EMBER & \small Full \\ \hline
500      & 0\% & 0\% & 0\% & 4\% & 0\% & 0\% & 1\% & 13\% & 13\% \\ \hline
2,000     & 0\% & 0\% & 0\% & 5\% & 1\% & 0\% & 2\% & 18\% & 30\% \\ \hline
5,000     & 0\% & 0\% & 0\% & 6\% & 2\% & 1\% & 2\% & 26\% & 52\% \\ \hline
10,000    & 0\% & 0\% & 0\% & 9\% & 2\% & 1\% & 1\% & 33\% & 71\% \\ \hline
\end{tabular}
\caption{Success Rate of the Append attacks for increased leverage on the Mini, EMBER and Full datasets.}
\label{table:srappends}
\end{table*}

\section{Results}
\label{sec:results}

Here, we evaluate the attacks described in the previous section in the same adversarial settings using models trained on the Mini, EMBER and Full datasets.
Our evaluation seeks to answer the following questions:
\begin{itemize}
  \item How do existing attacks generalize to classifiers trained on larger datasets?
  \item How vulnerable is a robust MalConv architecture to adversarial samples?
  \item Are slack-based attacks more effective than append attacks? 
  \item Are single-step adversarial samples transferable across models? 
\end{itemize}

In an attempt to reproduce prior work, we select candidate instances from the test set set if they have a file size smaller than 990,000 bytes and are correctly classified as malware by the victim. 
We randomly pick 400 candidates and measure the effectiveness of the attacks using the Success Rate (SR): the percentage of adversarial samples that successfully evaded detection.

\subsection{Append Attacks.}
We evaluate the append-based attacks on the Mini, EMBER and the Full datasets by varying the number of appended bytes, and summarize the results in Table~\ref{table:srappends}.
The Random Append attack fails on all three models, regardless of the number of appended bytes.
This result is in line with our expectations, demonstrating that the MalConv model is immune to random noise and that the input size is not among the learned features.
However, our results do not reinforce previously reported success rates of up to 15\% by Kolosnjaji et al.~\cite{kolosnjaji2018adversarial}.

The SR of the Benign Append attack seems to progressively increase with the number of added bytes on the Mini dataset, but fails to show the same behavior on the EMBER and Full datasets. 
Conversely, in the FGM Append attack we observe that the attack fails on the Mini dataset, while reaching up to 33\% SR on EMBER and 71\% SR on the Full datasets.  
This paradoxical behavior highlights the importance of large, robust datasets in evaluating adversarial attacks. 
One reason for the discrepancy in attack behaviors is that the MalConv model trained using the Mini dataset (modeled after the dataset used by Kolosnjaji et al.) has a severe overfitting problem. 
In particular, the success of appending specific benign byte sequences from the Mini dataset could be indicative of poor generalizability and this is further supported by the disconnect between the model's capacity and the number of samples in the Mini dataset. 
When we consider the single-step FGM Attack's success on the EMBER and Full datasets, and its failure on the Mini dataset, we believe these results can also be explained by poor generalizability in the Mini model; the single gradient evaluation does not provide enough information for the sequence of byte changes made in the attack. 
Recomputing the gradient after each individual byte change is expected to result in a higher attack success rate.
Finally, we also observe a large discrepancy between the SR on the EMBER and Full models, which counterintuitively highlights the model trained on a larger dataset as being more vulnerable. 
The results reveal an interesting property of single-step gradient-based atttacks: with more training data, the model encodes more sequential information and a single gradient evaluation becomes more beneficial for the attack. 
Conversely, updating the bytes independently of one another on the less robust model is less likely to succeed. 

Aside from the methodological issues surrounding dataset size and composition, our results also show that even a robustly trained MalConv classifier is vulnerable to append attacks when given a sufficiently large degree of freedom.
Indeed, the architecture uses 500 byte convolutional kernels with a stride size of 500 and a single max pool layer for the entire file, which means that not only is it looking at a limited set of relatively coarse features, but it also selects the best 128 activations locations irrespective of location.  That is, once a sufficiently large number of appended bytes are added in the FGM attack, they quickly replace legitimate features from the original binary in the max pool operation. 
Therefore, the architecture does not encode positional information, which is a significant vulnerability that we demonstrate can be exploited.

Additionally, we implemented the Gradient Append attack proposed by Kolosnjaji et al., but failed to reproduce the reported results.
We aimed to follow the original description, with one difference: our implementation, in line with the original MalConv architecture, uses a special token for padding, while Kolosnjaji et al. use the byte value $0$ instead.
We evaluated our implementation under the same settings as the other attacks, but none of the generated adversarial samples were successful.
One limitation of the Gradient Append attack that we identified is the necessity to update the value of each appended byte at each iteration.
However, different byte indexes might converge to their optimal value after a varying number of iterations.
Therefore, successive and unnecessary updates may even lead to divergence of some of the byte values.
Indeed, empirically investigating individual byte updates across iterations revealed an interesting oscillating pattern, where some bytes receive the same sequence of byte values cyclically in later iterations. 

\begin{figure*}[t!]
\centering
\begin{subfigure}{.3\textwidth}
  \centering
  \includegraphics[width=0.99\textwidth]{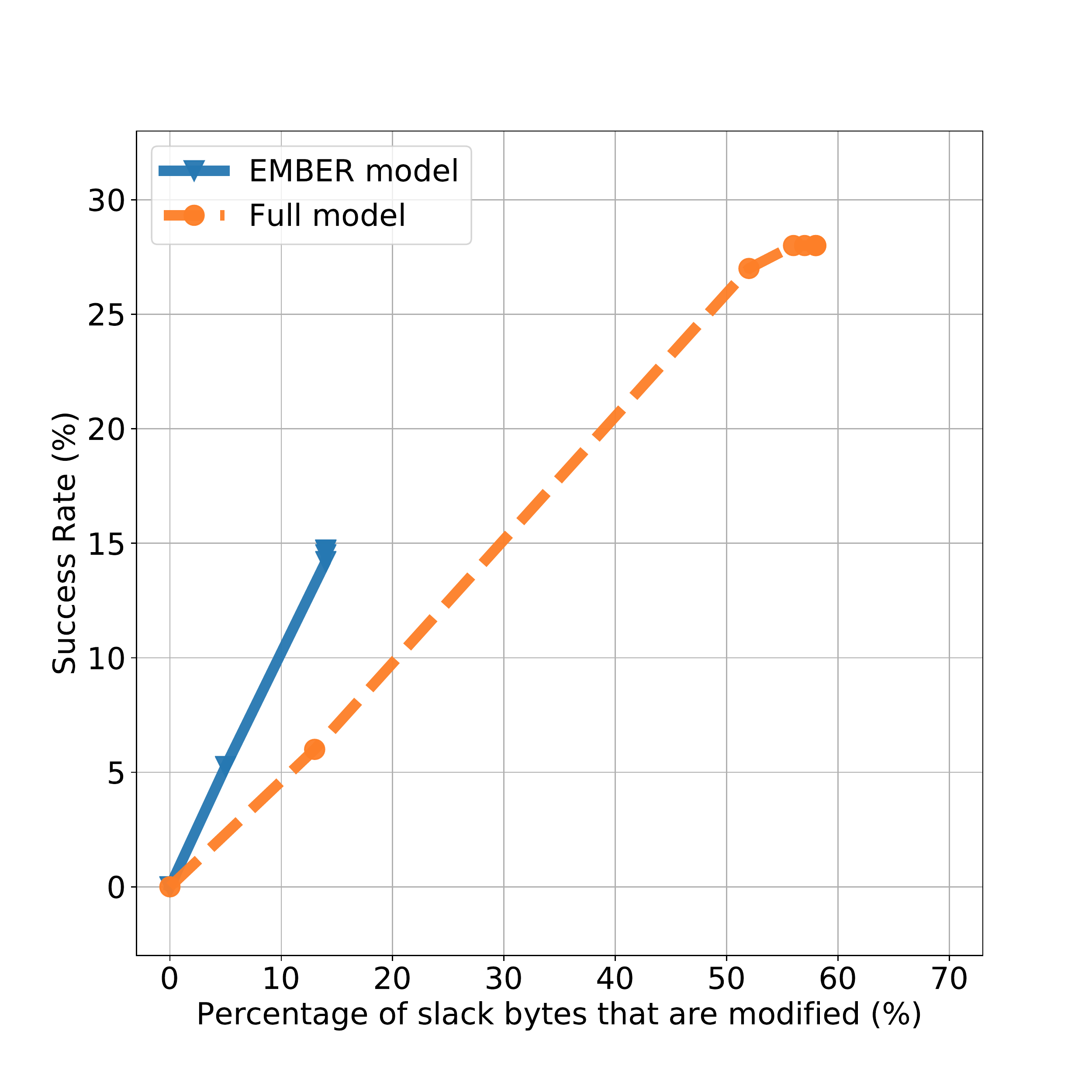}
  \caption{Slack FGM attack SR for increasing $\epsilon$}
  \label{fig:srnumbytesappend_sr}
\end{subfigure}\hfill
\begin{subfigure}{.3\textwidth}
  \centering
  \includegraphics[width=0.99\textwidth]{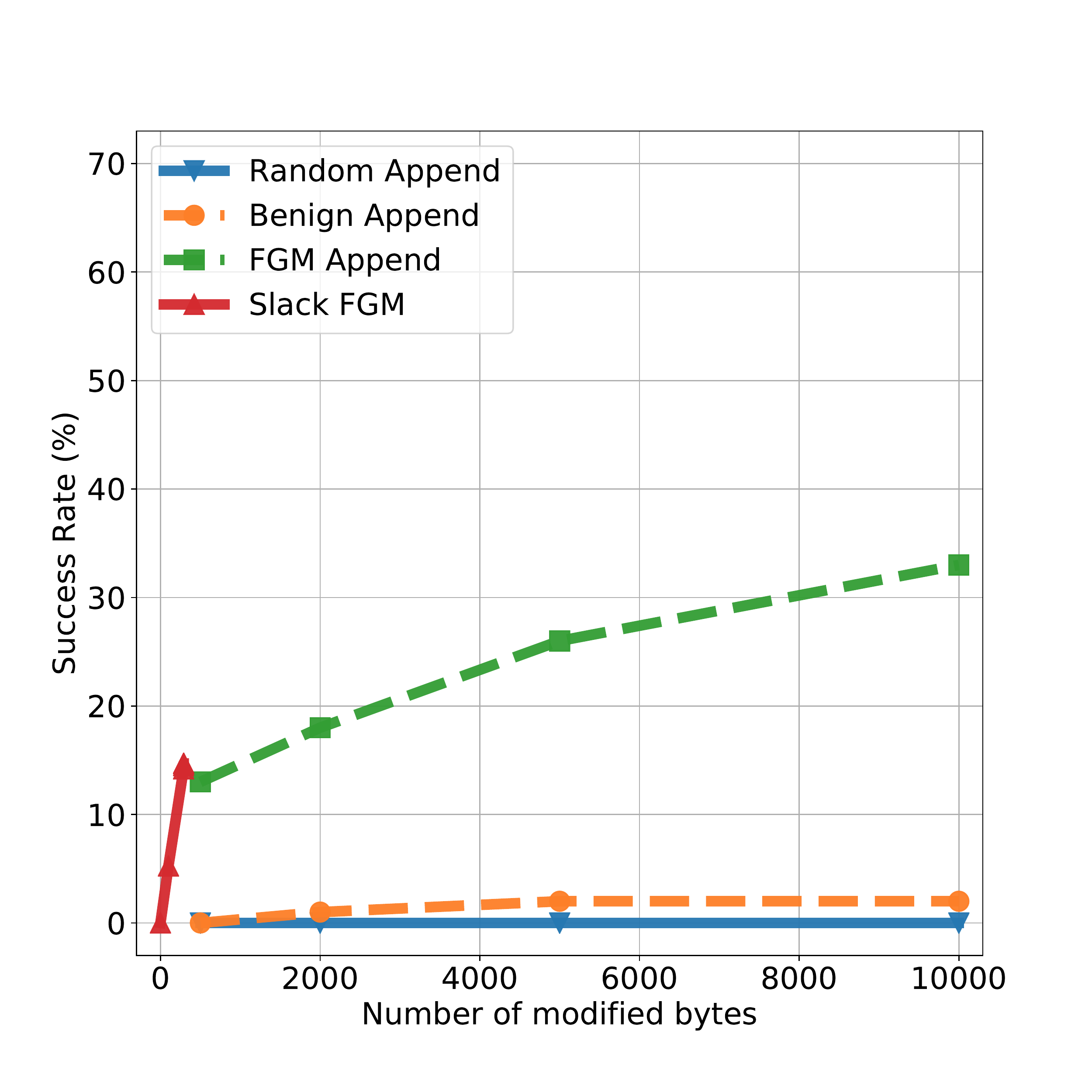}
  \caption{SR for EMBER Model}
  \label{fig:srnumbytesappend_ember}
\end{subfigure}\hfill
\begin{subfigure}{.3\textwidth}
  \centering
  \includegraphics[width=0.99\textwidth]{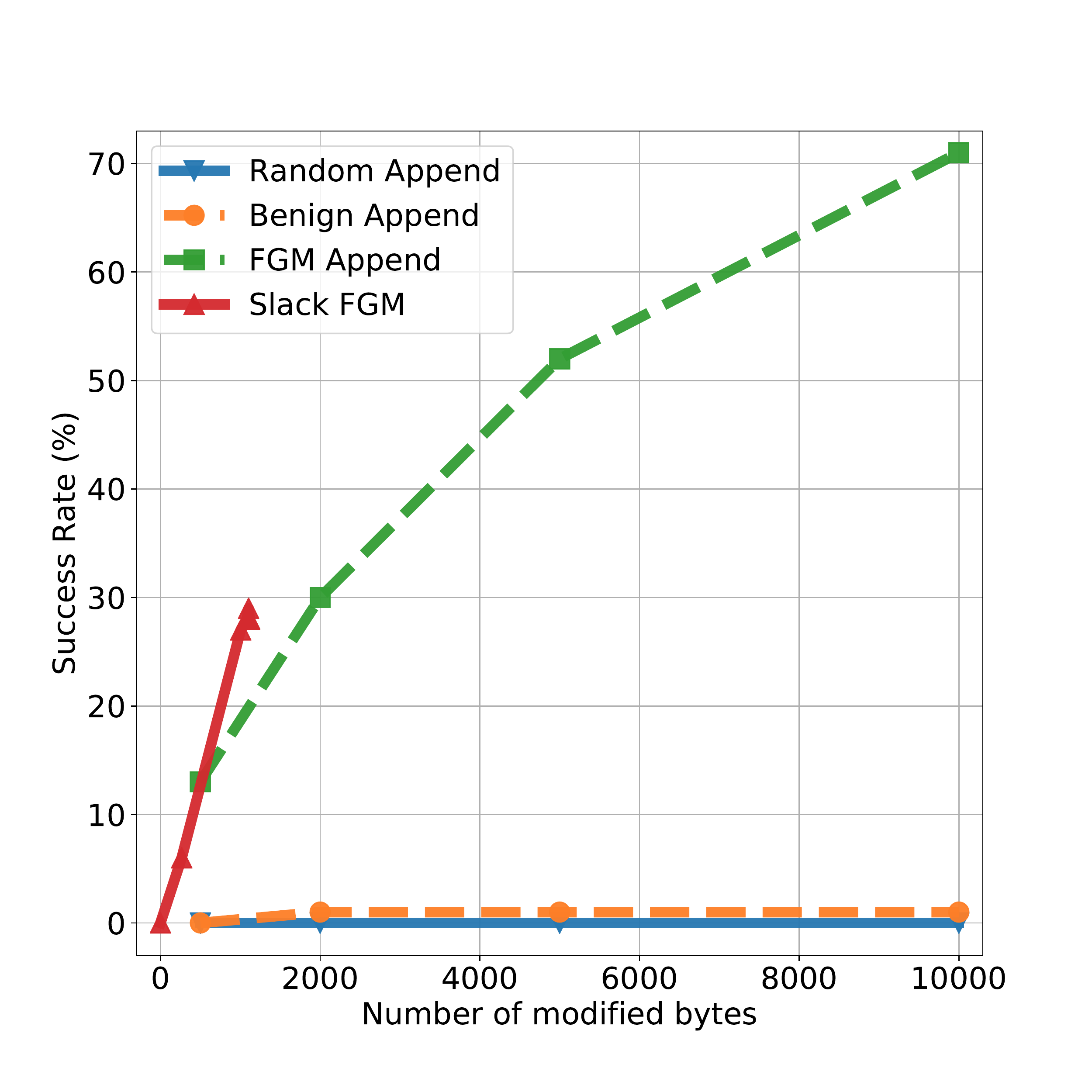}
  \caption{SR for Full Model}
  \label{fig:srnumbytesappend_full}
\end{subfigure}
\caption{Evaluation of the Slack FGM attack on the EMBER and Full models.}
\label{fig:srnumbytesappend}
\end{figure*}

\subsection{Slack Attacks.} 
We evaluate the Slack FGM attack over the EMBER and Full datasets for the same experimental settings as above.  
In order to control the amount of adversarial noise added in the slack bytes, we use the $\epsilon$ parameter to define an $L_2$ ball around the original byte value in the embedding space.  
Only those values provided by the FGM attack that fall within the $\epsilon$ ball are considered for the slack attack, otherwise the original byte value will remain.
As illustrated in Figure~\ref{fig:srnumbytesappend_sr}, by varying $\epsilon$ we control the percentage of available slack bytes that are modified. 
The upper bound for the SR is 15\% on EMBER for an attack where 14\% (291/2103) slack bytes were modified on average, while on Full we achieve 28\% SR for 58\% (1117/1930).
While the attack is more successful against Full than EMBER, it also succeeds in modifying a proportionally larger number of bytes. 
We observe that the EMBER model returns very small gradient values for the slack bytes, indicating that their importance for classifying the target is low. 
The results also reinforce our hypothesis about the single gradient evaluation on the FGM Append attack.

In order to compare Slack FGM with the append attacks, in Figures~\ref{fig:srnumbytesappend_ember} and~\ref{fig:srnumbytesappend_full} we plot the SR as a function of the number of modified bytes.
The results show that, while the FGM Append attack could achieve a higher SR, it also requires a much larger number of byte modifications. 
On EMBER, Slack FGM modifies 291 bytes on average, corresponding to a SR for which FGM Append requires approximately 500 bytes.
On Full, the attack achieves a SR of 27\% for an average of 1005 modified bytes, while the SR of the FGM Append lies around 20\% for the same setting.
The results confirm our initial intuition that the coarse nature of MalConv's features requires consideration of the surrounding contextual bytes within the convolutional window.  
In the slack attack, we make use of existing contextual bytes to amplify the power of our FGM attack without having to generate a full 500-byte convolutional window using appended bytes.

\subsection{Attack Transferability.} 
We further analyze the transferability of attack samples generated for one (source) model against another (target). 
We run two experiments with EMBER and Full alternately acting as source and target, and evaluate FGM Append and Slack FGM attacks on samples that successfully evade the source model and for which the original (pre-attack) sample is correctly classified by the target model. 
At most 2/400 samples evade the target model for each set of experiments, indicating that these \textit{single-step samples are not transferable between models.} 
The findings are not in line with prior observations on adversarial examples for image classification, where single-step samples were found to successfully transfer across models~\cite{kurakin2016adversarial}.
Nevertheless, we leave a systematic transferability analysis of other embedding mappings and stronger iterative attacks for future work.

\section{Related Work}
\label{sec:related}
The work by Barreno et al.~\cite{barreno2010security} was among the first to systematize attack vectors against machine learning, where they distinguished evasion as a type of test-time attack.  Since then, several evasion attacks have been proposed against malware detectors.  Many of these attacks focus on additive techniques for evasion, where new capabilities or features are added to cause misclassification.  For instance, Biggio et al.~\cite{biggio2013evasion} use a gradient-based approach to evade detection by adding new features to PDFs, while Grosse et al. ~\cite{grosse2017adversarial} and Hu et al.~\cite{hu2017black} add new API calls to evade detection. Al-Dujaili et al.~\cite{huang2018adversarial} propose an adversarial training framework against these additive attacks. More recently, Anderson et al.~\cite{anderson2018learning} used reinforcement learning to evade detectors by selecting from a pre-defined list of semantics-preserving transformations.  Similarly, Xu et al.~\cite{xu2016automatically} propose a genetic algorithm for manipulating PDFs while maintaining necessary syntax.  Closest to our work are the gradient-based attacks by Kolosnjaji et al.~\cite{kolosnjaji2018adversarial} and Kreuk et al. \cite{kreuk2018deceiving} against the MalConv architecture.
By contrast, our attacks are intended to highlight trade-offs between the append and slack strategies, and to test the robustness of the MalConv architecture when trained on production-scale datasets.
Additionally, to our knowledge, the transferability of single-step adversarial attacks on malware has not been previously studied despite prior work that suggests it is best suited for mounting black-box attacks~\cite{kurakin2016adversarial}.

\section{Conclusion}
\label{sec:conclusion}
In this paper, we explored the space of adversarial examples against deep learning-based malware detectors.
Our experiments indicate that the effectiveness of adversarial attacks on models trained using small datasets does not always generalize to robust models.
We also observe that the MalConv architecture does not encode positional information about the input features and is therefore vulnerable to append-based attacks.
Finally, our attacks highlight the threat of adversarial examples as an alternative to evasion techniques such as runtime packing. 


\section*{Acknowledgments}
We thank Jon Erickson for helpful discussions with regard to slack attack methods and the anonymous reviewers for their constructive feedback.

\bibliographystyle{IEEEtran}
\bibliography{IEEEabrv,aml}

\end{document}